\def\year{2019}\relax
\documentclass[letterpaper]{article} 
\usepackage{aaai19}  
\usepackage{times}  
\usepackage{helvet}  
\usepackage{courier}  
\usepackage{url}  
\usepackage{graphicx}  
\usepackage{url}
\usepackage{booktabs}
\usepackage{multirow}
\usepackage{colortbl}
\usepackage{latexsym}
\usepackage{amsmath}
\usepackage{amssymb}
\usepackage{color}
\usepackage{float}
\usepackage{tabularx}
\usepackage{subfig}
\usepackage{multirow}
\usepackage{latexsym}
\usepackage{amsmath}
\usepackage{amssymb}
\usepackage{colortbl}
\usepackage{array}
\usepackage{bm}
\usepackage{booktabs}
\usepackage{pgfplots}

\usepackage{algorithm, algorithmic}

\renewcommand{\algorithmiccomment}[1]{\bgroup\hfill//~#1\egroup}

\newcommand{\citet}[1]
{\citeauthor{#1}~\shortcite{#1}}
\newcommand{\citep}{\cite}

\def\ba{\mathbf{a}}

\def\be{\mathbf{e}}

\def\bh{\mathbf{h}}

\def\bs{\mathbf{s}}

\def\cJ{\mathcal{J}}

\def\cM{\mathcal{M}}

\def\cR{\mathcal{R}}

\allowdisplaybreaks

\DeclareMathOperator*{\softmax}{softmax}

\allowdisplaybreaks
\newenvironment{itemize*}%
  {\begin{itemize}%
    \setlength{\itemsep}{1pt}%
    \setlength{\parskip}{1pt}}%
  {\end{itemize}}
  \newenvironment{enumerate*}%
  {\begin{enumerate}%
    \setlength{\itemsep}{1pt}%
    \setlength{\parskip}{1pt}}%
  {\end{enumerate}}
\usepackage{url}

\begin{document}
\title{Exploring Shared Structures and Hierarchies for Multiple NLP Tasks}
\author{Junkun Chen\thanks{\quad Equal contribution}, Kaiyu Chen\footnotemark[1], Xinchi Chen, Xipeng Qiu\thanks{\quad Corresponding Author}, Xuanjing Huang\\
Shanghai Key Laboratory of Intelligent Information Processing, Fudan University\\
School of Computer Science, Fudan University\\
  {\tt \{jkchen16, kychen15, xinchichen13, xpqiu, xjhuang\}@fudan.edu.cn} \\
   \\}
\maketitle
\begin{abstract}
Designing shared neural architecture plays an important role in multi-task learning. The challenge is that finding an optimal sharing scheme relies heavily on the expert knowledge and is not scalable to a large number of diverse tasks.
Inspired by the promising work of neural architecture search (NAS), we apply reinforcement learning to automatically find possible shared architecture for multi-task learning.
Specifically, we use a controller to select from a set of shareable modules and assemble a task-specific architecture, and repeat the same procedure for other tasks.
The controller is trained with reinforcement learning to maximize the expected accuracies for all tasks. We conduct extensive experiments on two types of tasks, text classification and sequence labeling, which demonstrate the benefits of our approach.
\end{abstract}

\section{Introduction}


Multi-task Learning (MTL) is an essential technique to improve the performance of a single task by sharing the experiences of other related tasks. A fundamental aspect of MTL is designing the shared schemes. This is a challenging ``meta-task'' of itself, often requires expert knowledge of the various NLP tasks at hand.

\begin{figure}[t]\small
\centering
\includegraphics[width=1\linewidth]{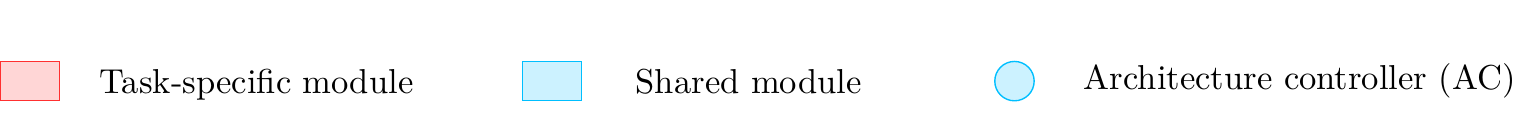}
\\
\hspace{-2.7em}
\subfloat[Hard]{
  \includegraphics[width=0.32\linewidth]{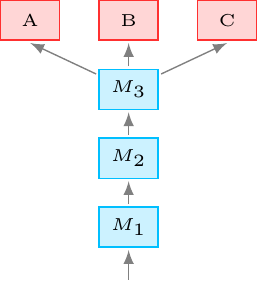}\label{fig:hard}
  }
 \hspace{1.5em}
\subfloat[Soft]{
  \includegraphics[width=0.32\linewidth]{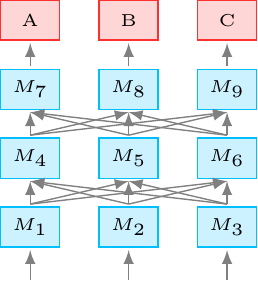}\label{fig:soft}
  }
  \hspace{7em}
  \\
\subfloat[Hierarchical]{
  \includegraphics[width=0.32\linewidth]{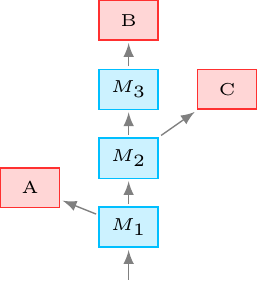}\label{fig:hierarchical}
  }
  \hspace{1em}
\subfloat[This work]{
  \includegraphics[width=0.45\linewidth]{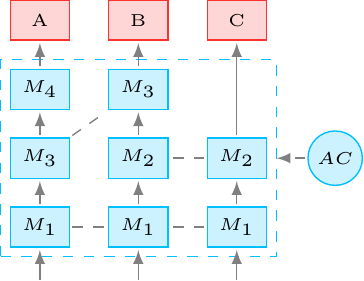}\label{fig:ours}
  }

  \caption{Sharing schemes of multi-task learning. Modules connected by dashed lines in (\ref{fig:ours}) are identical, selected from a shared module pool. }\label{fig:Exsit}
\end{figure}

Figure \ref{fig:Exsit} shows, broadly, the categories of existing MTL schemes:
\begin{enumerate*}
  \item \textbf{Hard-sharing}: All tasks share a common set of modules before fanning out to their privates ``heads''. 
  This is a basic shared strategy and an efficient approach to fight overfitting \citep{baxter1997bayesian}.

  \item \textbf{Soft-sharing}: In this scheme, modules of each task are \textit{shareable} and can contribute to different tasks. Consequently, the input to a task's certain module is a weighted sum of the outputs from depending modules of all tasks; the weights can differ across tasks~\citep{misra2016cross}. This scheme is much more flexible.

  \item \textbf{Hierarchical-sharing} When the tasks are related but belong to different stages in a pipeline or general data flow graph, the sharing scheme should have a hierarchical structure, where lower level tasks branching off earlier to feed to their private heads~\citep{sogaard2016deep}.
  This scheme is very suitable for NLP tasks since most of NLP tasks are related and there is an implicit hierarchy among their representations. A well-designed hierarchical sharing scheme can boost the performance of all tasks significantly \citep{hashimoto2017joint}.
\end{enumerate*}

\begin{figure*}[h]\centering
 \includegraphics[width=0.75\linewidth]{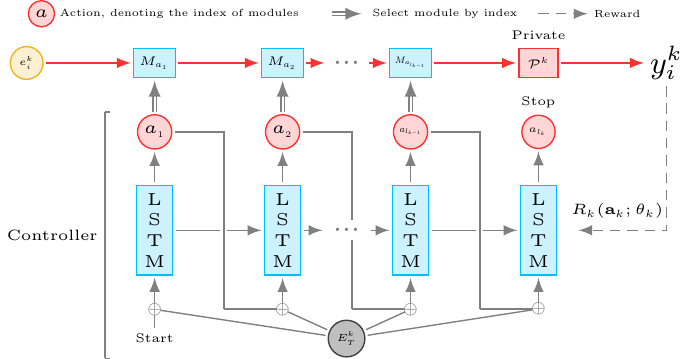}
 \caption{The architecture of proposed model. The blue modules are shared between different tasks. The RNN-based controller consecutively selects M$_{a_{1}}$, M$_{a_{2}}$, $\cdots$,  M$_{a_{l_k}}$ for task $k$. $a_t$ is the action of step $t$ generated by controller, indicating the index of selected modules. Finally, the controller generates a $Stop$ action, terminating the selection process. $\mathcal{P}^{k}$ denotes to the private module of task $k$. $\oplus$ is a concatenation operation. The reward of the controller is the performance of the task $k$ under the selected architecture of the shared layers. }\label{fig:rl_mtl}
\end{figure*}

Despite their success, there are still two challenging issues.

The first issue is that judging relatedness of two tasks is not easy, and it is even more difficult of when number of tasks scale. 
For instance, consider three classification tasks for movie reviews, product reviews, and spam detection. A preferred option is sharing information between movie reviews and product reviews classification, but not spam detection. However, it is hard to explore all configurations when number of tasks become large.
An improper sharing scheme design may cause ``negative transfer'', a severe problem in MTL and transfer learning. Besides, it suffers from a problem of ``architecture engineering''. Once datasets alter, we need to redesign and explore a new sharing scheme. This is taxing even for experts. 


The second issue is how to design a hierarchical-sharing scheme for multi-level tasks. For instance, many tasks in NLP are related but not belonging to the same level, such as part-of-speech (POS) tagging, named entity recognition (NER), semantic role labeling (SRL), etc. Thus, the hierarchical-sharing scheme should be adopted. However, it is non-trivial to choose the proper sharing structure manually especially facing a large set of tasks according to the expert knowledge. It also suffers from the problems of ``negative transfer'' and ``architecture engineering''.

To address the above two challenging issues for MTL, in this paper, we learn a controller by reinforcement learning to automatically design the shared neural architecture for a group of tasks, inspired by a recent promising work, neural architecture search (NAS)~\citep{zoph2016neural}. Specifically, there is a pool of sharable modules, where one or multiple modules can be integrated into an end-to-end task-specific network. As shown in Figure \ref{fig:ours}, for each task, the controller chooses one or multiple modules from sharable module pool, stacks them in proper order, and then integrates them into a task-specific network. The controller is trained to maximize the expected accuracies of its designed architectures for all the tasks in MTL. Additionally, we also propose a novel training strategy, namely \textit{simultaneous on-line update strategy}, which  accelerates the training procedure.

The benefits of our approach could be summarized as follows.
\begin{itemize*}

\item Since not all the sharable modules are used in a task-specific network, 
the related tasks can share overlapping modules, whereas the unrelated tasks keep their private modules.
\item The controller can stack the chosen modules in various orders. Therefore its designed architecture can handle the multi-level tasks. One shared module can posit at low-layer for one high-level task, but at high-layer for another low-level task, therefore a hierarchical-sharing structure is implicitly constructed.
\item Overall, our approach provides a flexible solution to the problem of designing shared architecture for MTL and can handle complicated correlation and dependency relationship among multiple tasks.
\end{itemize*}


\begin{algorithm*}[t]\small
\begin{algorithmic}[1]
\REPEAT
  \FOR {each batch}
        \FOR {each task $k$}

            \STATE Drawn $\ba_1,\ba_2,\cdots,\ba_N$ $\sim$ $\pi_\phi(E_T(k))$ \COMMENT{Sample $N$ structures with policy $\pi_\phi(E_T(k))$}
            \FOR {each $\ba \in \{\ba_1,\ba_2,\cdots,\ba_N\}$}
 \STATE Update $\theta \leftarrow \theta+\alpha\nabla_\theta R_k(\ba;\theta)$ \COMMENT{Update the parameters of shared modules}
 \ENDFOR
            \STATE Calculate $N$ rewards $\cR_k = \{R_k(\ba_1), R_k(\ba_2), \cdots, R_k(\ba_N)\}$ \COMMENT{$R_k$ is calculated by Eq (\ref{eq:rr})}
          \STATE Normalize rewards $\bar{R}_k$ = $\softmax(\cR_k, \tau)$ \COMMENT{Eq (\ref{eq:normalize})}
          \STATE Update $\phi \leftarrow \phi+\alpha\nabla_\phi \bar{R}_k$ \COMMENT{Update the parameters of controller}



        \ENDFOR
  \ENDFOR
\UNTIL{$\theta$ and $\phi$ convergence}
\end{algorithmic}
\caption{\textbf{Training with Simultaneous On-line Update Strategy}}
\label{alg:rl_mtl}
\end{algorithm*}

\section{Model}
Usually, a multi-task framework contains shared layers and task-specific layers. The proposed model gives a mechanism for automatically learning the structure of the shared layers for each task. The model contains $L$ shared modules: $\cM$ = \{M$_1$, M$_2$, $\cdots$, M$_L$\}, and a RNN-based controller will select modules for each task. Figure \ref{fig:rl_mtl} gives an illustration.

\subsection{Training Controller with REINFORCE}
The structure of the shared layers for task $k$ is selected by a RNN-based controller (a list of actions $\ba_k = \{a_1, a_2, \cdots, a_{l_k}\}$), which takes the task embedding of $E_T(k)$ of task $k$ as the input. The chosen layers stacked one by one form the shared part of the model for the specific task $k$. And performance under the chosen structure gives the reward feedback $R_k$ for the controller. Thus, we can use reinforcement learning to train the controller. Concretely, our objective is to find the optimal with policy $\pi_\phi(E_T(k), \ba_k)$ by maximizing the expected reward $\cJ_k(\phi)$:
\begin{align}
&\cJ_k(\phi) =\mathbb{E}_{\pi_\phi(E_T(k), \ba_k)} [R_k(\ba_k;\theta_k)].
\label{eq:fir}
\end{align}
Since the $R_k(\cdot)$ is non-differentiable, we need to use a policy gradient method to optimize $\phi$, thus the derivative of $\cJ_k(\phi)$ is:

\begin{equation}
\begin{aligned}
& \nabla_\phi \cJ_k(\phi) = \\
\sum_{t=1}^{l_k}\mathbb{E} [\nabla_\phi \log & \pi_\phi(E_T(k), a_t) R_k(\ba_k;\theta_k)].
\end{aligned}
\end{equation}

In this paper, we use LSTM to model $\pi_\phi(E_T(k), a_t)$, where $a_t$ indicates a selection among $L$ shared modules $\cM$.

\subsection{Reward Function}
The reward is the performance of the predicted architecture on task $k$. Specifically, given training set $\{x_i^k, y_i^k\}_{i=1}^{N_k}$ of task $k$, we firstly map the input sentence $x_i^k$ to $\be_{i}^k$ with word embeddings. Assuming that the predicted actions are $\ba_k = \{a_1, a_2, \cdots, a_{l_k}\}$ and the selected modules are $\cM_{\ba_k} = \{M_{a_1}, M_{a_2}, \cdots, M_{a_{l_k}}\}$, the reward $R_k(\ba_k;\theta)$ is:
\begin{gather}
\be_{i}^k \xrightarrow{\cM_{\ba_k}} \bs_i^k \xrightarrow{\mathcal{P}^{k}} p(y_i^k|x_i^k) \\
R_k(\ba_k;\theta) = \frac{1}{N_k}\sum_{i=1}^{N_k} \log p(y_i^k|x_i^k).\label{eq:rr}
\end{gather}
In this paper, modules are Bi-LSTM layers.

\section{Training with Simultaneous On-line Update Strategy}
Our model is composed of two parts. The learning modules focus on processing textual data for each task given the architecture. The controller aims to find a better structure for learning modules to achieve higher performance.

These two parts are trained together with our on-line update strategy demonstrated in Algorithm \ref{alg:rl_mtl}, which means we simultaneously train these two parts at each batch. It is different from previous work, where the controller updates itself only when the multi-task learning procedure completed. A significant advantage of simultaneous training strategy is its high efficiency compared with previous training strategy. 

In implementing on-line update strategy, we have three major components as follows.

\paragraph{Multi-sampling of Architecture Controller}
As the amplitudes and distributions of rewards of each task are probably distinct, rewards can only be privately used for individual tasks, causing a problem of insufficient rewards. To solve the problem and help the controller distinguish the performance of different architectures, we sample $N$ times of rewards for each single training instance. These rewards are subsequently used for training the controller after reward normalization.

\paragraph{Reward Normalization}
Given $N$ rewards $\cR_k = \{R_k^{(1)}, R_k^{(2)}, \cdots, R_k^{(N)}\}$ for task $k$, $\bar{R}_{k}$ could be finally derived by normalizing these rewards with a parameterized $\softmax$ function \citep{norouzi2016reward}:
\begin{equation}
\bar{R}_{k}  = {\frac {\exp(R_{k}^{(i)}/\tau)}{\sum _{j=1}^{N}\exp(R_{k}^{(j)}/ \tau )}},\label{eq:normalize}
\end{equation}
where $\tau$ is a hyper-parameter that adjusts the variance of $\bar{R}_{k}$. This normalization method maps rewards to [0, 1], will stabilize the training process.


\section{Experiment}
In this section, we evaluate our model on two types of tasks, text classification, and sequence labeling.
\subsection{Exp-I: Text Classification}
Firstly, we apply our model to a text classification dataset whose tasks are separated by domains. With this dataset, we can check if our model can effectively use shared information across domains.

\begin{table*}[t]\setlength{\tabcolsep}{15pt}\small
\centering
\begin{tabular}{c|cc|ccccc}
\midrule
                        & \multicolumn{2}{c|}{Single-Task} & \multicolumn{5}{c}{Multi-Task} \\ 
\multirow{-2}{*}{Tasks} & LSTM  &Bi-LSTM & FS     & SSP   & PSP  & CS & Our  \\ \midrule
Books                   & 80.3  & 82.0 & 84.5   & 87.0  & 84.7 & 84.0 & 87.3  \\ 
Electronics             & 78.5  & 80.0 & 86.7   & 86.5  & 85.5 & 85.5 & 88.3  \\ 
DVD                     & 78.3  & 84.0 & 86.5   & 86.3  & 86.0 & 86.5 & 86.0  \\ 
Kitchen                 & 81.7  & 83.5 & 86.0   & 88.7  & 88.0 & 87.3 & 87.5  \\ 
Apparel                 & 83.0  & 85.7 & 85.7   & 86.3  & 87.5 & 87.0 & 88.7  \\ 
Camera                  & 87.3  & 88.3 & 86.7   & 88.0  & 87.7 & 86.3 & 90.0  \\ 
Health                  & 84.5  & 85.3 & 89.3   & 90.0  & 89.0 & 90.0 & 90.3  \\ 
Music                   & 77.5  & 78.7 & 82.5   & 81.0  & 81.3 & 84.0 & 84.7  \\ 
Toys                    & 84.0  & 85.7 & 88.5   & 87.7  & 88.5 & 87.5 & 89.7  \\ 
Video                   & 80.3  & 83.7 & 86.7   & 85.5  & 84.5 & 85.7 & 86.0  \\ 
Baby                    & 85.7  & 85.7 & 88.0   & 87.7  & 87.3 & 88.3 & 89.0  \\ 
Magazines               & 92.3  & 92.5 & 91.5   & 93.0  & 93.3 & 91.0 & 90.3  \\ 
Software                & 84.5  & 86.7 & 87.0   & 88.3  & 89.7 & 87.5 & 91.0  \\ 
Sports                  & 79.3  & 81.0 & 85.5   & 86.5  & 86.7 & 87.7 & 88.7  \\ 
IMDB                    & 79.5  & 83.7 & 82.3   & 85.3  & 84.5 & 82.5 & 84.7  \\ 
MR                      & 73.5  & 73.7 & 71.0   & 72.7  & 72.5 & 73.3 & 76.3  \\ \midrule
\rowcolor[HTML]{C0C0C0}
Avg                     & 81.9  & 83.0 & 85.5   & 86.3   & 86.1 & 85.9 & \textbf{87.4}  \\ \bottomrule
\end{tabular}
\caption{The results of text classification experiment.}
\label{tab:result_textclass}
\end{table*}

\paragraph{Data}

We use a dataset of 16 tasks. The dataset is a combination of two categories. The first category that makes up 14 tasks is product reviews, extracted from the dataset\footnote{\url{https://www.cs.jhu.edu/~mdredze/datasets/sentiment/}} constructed by \citep{blitzer2007biographies}. Reviews of each task come from different Amazon products domain, such as Books, DVDs, and so on. The second category making up the remaining 2 tasks is movie reviews, specifically IMDB\footnote{\url{https://www.cs.jhu.edu/~mdredze/datasets/sentiment/unprocessed.tar.gz}} conducted by \citep{maas2011learning}, and MR\footnote{\url{https://www.cs.cornell.edu/people/pabo/movie-review-data/}} from rotten tomato website, conducted by \citep{pang2005seeing}.

The dataset for each task contains 2000 samples, which partitioned randomly into training data, development data and testing data with the proportion of 70\%, 10\%, and 20\% respectively. Each sample has two class labels, either positive or negative.

\paragraph{Baseline Model}
We compare our proposed model with four multi-task models and one single-task model. These baseline models have the same Bi-LSTM modules with proposed model. They differ only in how we organize these modules.
\begin{itemize*}
\item \textbf{FS-MTL}: All tasks use a fully-shared module and a task-specific classifier.
\item \textbf{SSP-MTL}: This model is followed stack share-private scheme. There is a shared module that receives embed inputs of all tasks and takes its outputs as inputs for task-specific modules. The outputs of task-specific modules are then fed into the task-specific classifier.
\item \textbf{PSP-MTL}: This model is followed a parallel share-private scheme. Concatenate the outputs of a fully-shared module with the feature extracted from task-specific modules as inputs to the task-specific classifier.
\item \textbf{CS-MTL}: Cross Stitch multi-task model proposed by \citep{misra2016cross}. It's a soft-sharing approach, using units to model shared representations as a linear combination.
\item \textbf{Single Task}: Standard Bi-LSTM classifier that trains separately on each task.

\end{itemize*}

The classifier mentioned above is a linear layer taking the average of the text sequence as inputs.

\paragraph{Hyperparameters}
The networks are trained with backpropagation, and the gradient-based optimization is performed using the Adam update rule \citep{kingma2014adam}, with an early-stop parameter of 12 epochs.

As for the controller, the size of the RNN's hidden state is 50. Additionally, an exploration probability of sampling randomly is set to 0.2, and the scaling factor in reward normalization $\tau$ is $\frac{1}{30}$. The $N = 20$ modules are available, but the model tends to use less than $N = 20$ modules. The size of task embeddings $S$ is 15.

The word embeddings for all of the models are initialized
with the 100d GloVe vectors (6B token version) \citep{pennington2014glove}, while the size of hidden state $\bh$ in Bi-directional LSTMs is 50. Per-task Fine-tuning during training with shared parameters fixed is used to improve the performance. The mini-batch size is set to 64.

\begin{table}[t]\setlength{\tabcolsep}{4pt}\small
\centering
\begin{tabular}{c|rrr}
\toprule
Domains & \multicolumn{1}{c}{Train}   & \multicolumn{1}{c}{Dev}     & \multicolumn{1}{c}{Test}   \\ \midrule
\begin{tabular}[c]{@{}c@{}}Broadcast  Conversation (bc)\end{tabular} & 173,289  & 29,957   & 35,947  \\
\begin{tabular}[c]{@{}c@{}}Broadcast  News (bn)\end{tabular}         & 206,902  & 25,271   & 26,424  \\ 
Magazines (mz)                                                         & 164,217  & 15,421   & 17,874  \\ 
Newswire (nw)                                                          & 878,223  & 147,955  & 60,756  \\ 
Pivot Corpus (pc)                                                      & 297,049  & 25,206   & 25,883  \\ 
\begin{tabular}[c]{@{}c@{}}Telephone  Conversation (tc)\end{tabular} & 90,403   & 11,200   & 10,916  \\ 
Weblogs (wb)                                                           & 388,851  & 49,393   & 52,225  \\ \bottomrule
\end{tabular}
\caption{Statistics of tokens for each domain in OntoNote 5.0 dataset.} \label{tab:ontonote}
\end{table}

\paragraph{Result}

We evaluate all these models, and the classifying accuracy is demonstrated in Table \ref{tab:result_textclass}. Compared with the best baseline model, our model still achieves higher performance by 1.5 percent. With more observation of the final architecture selected by the controller, we find out that the controller groups similar tasks together, showing the ability of clustering, which will be further discussed in Section 5.


\begin{table*}[t]\setlength{\tabcolsep}{5pt}\small
\renewcommand{\arraystretch}{1.3}
\centering
\begin{tabular}{c|cccccc|cccccc}
\toprule

Domain & Task  & Bi-LSTM & PSP   & SSP   & CS    & Our   & Task & Bi-LSTM & PSP   & SSP   & CS    & Our   \\ \midrule
\multirow{2}{*}{bc}      & POS   & 89.56  & 94.21 & \textbf{94.67} & 94.17 & 94.61 & NER  & 92.15  & 94.26 & \textbf{94.50} & 93.03 & \textbf{94.50} \\
       & CHUNK & 85.60  & 89.20 & 89.81 & 86.24 & \textbf{91.19} & SRL  & 97.58  & 98.07 & 98.11 & 97.80 & \textbf{98.21} \\\midrule
\multirow{2}{*}{bn}     & POS   & 92.26  & 94.67 & 95.17 & 92.49 & \textbf{95.28} & NER  & 88.46  & 93.19 & 93.11 & 89.70 & \textbf{93.68} \\
       & CHUNK & 87.23  & 90.80 & 91.06 & 86.02 & \textbf{91.34} & SRL  & 95.56  & \textbf{97.94} & 97.71 & 97.66 & 97.35 \\\midrule
\multirow{2}{*}{mz}     & POS   & 79.77  & 91.22 & 92.80 & 84.61 & \textbf{92.88} & NER  & 89.69  & 92.09 & 93.22 & 91.23 & \textbf{93.37} \\
       & CHUNK & 83.79  & 89.85 & 90.36 & 81.81 & \textbf{91.04} & SRL  & 97.75  & 97.52 & 97.86 & 97.96 & \textbf{98.16} \\\midrule
\multirow{2}{*}{nw}     & POS   & 94.35  & 95.11 & 94.80 & \textbf{96.23} & 95.02 & NER  & 93.27  & 94.30 & 93.62 & \textbf{95.48} & 93.94 \\       & CHUNK & 90.51  & 92.19 & 91.00 & 92.25 & \textbf{92.29} & SRL  & 96.45  & 97.05 & 96.73 & 96.83 & \textbf{97.81} \\\midrule
\multirow{2}{*}{wb}     & POS   & 89.40  & 92.69 & \textbf{93.24} & 90.08 & 92.71 & NER  & 96.20  & 96.48 & 96.51 & \textbf{98.17} & 96.58 \\
       & CHUNK & 87.62  & 91.39 & 91.56 & 88.00 & \textbf{92.52} & SRL  & 95.13  & 95.17 & 96.17 & 95.84 & \textbf{96.36} \\\hline
\multirow{2}{*}{tc}     & POS   & 92.75  & 93.88 & 94.63 & 94.24 & \textbf{94.99} & NER  & 95.25  & 95.32 & 95.37 & \textbf{97.21} & 95.95 \\
       & CHUNK & 88.05  & 90.41 & 90.23 & 87.35 & \textbf{91.38} & SRL  & 98.32  & 98.80 & 98.84 & 98.81 & \textbf{99.05} \\\midrule
\multirow{2}{*}{pt}     & POS   & 97.26  & 97.73 & 97.54 & \textbf{98.81} & 97.68 & NER  & /      & /     & /     & /     & /     \\
       & CHUNK & 93.86  & 95.33 & 94.45 & 95.47 & \textbf{95.91} & SRL  & 96.38  & 96.35 & 96.36 & 96.20 & \textbf{97.42} \\\midrule
\rowcolor[HTML]{C0C0C0}
   & POS   & 90.76  & 94.22 & 94.69 & 92.95 & \textbf{94.74} & NER  & 92.50  & 94.27 & 94.39 & 94.14 & \textbf{94.67} \\
\rowcolor[HTML]{C0C0C0}

\multirow{-2}{*}{Avg} & CHUNK & 88.09  & 91.31 & 91.21 & 88.16 & \textbf{92.24} & SRL  & 96.74  & 97.27 & 97.40 & 97.30 & \textbf{97.77} \\\bottomrule
\end{tabular}
\caption{Resultant accuracy for the sequence labeling dataset, regarding every combination of (domain, problem) as tasks, accumulating to 7 * 4 - 1 = 27 separate tasks (accurately 27 tasks, since pt has no NER annotation).}
\label{tab:ontonotes_sep}
\end{table*}

\begin{table}[t]\setlength{\tabcolsep}{15pt}\small
\renewcommand{\arraystretch}{1.5}
\centering
\begin{tabular}{c|c|c}
\hline
      & \begin{tabular}[c]{@{}c@{}}Unite \\ Domains\end{tabular} & \begin{tabular}[c]{@{}c@{}}Separated \\ Domains\end{tabular} \\ \hline
POS   & 94.23                                                    & \textbf{94.74}                                                        \\
CHUNK & 91.86                                                    & \textbf{92.24}                                                        \\
NER   & 94.58                                                    & \textbf{94.67}                                                        \\
SRL   & 97.12                                                    & \textbf{97.77}                                                        \\ \hline
\rowcolor[HTML]{C0C0C0}
Avg.  & 94.45                                                    & \textbf{94.85}                                                        \\ \hline
\end{tabular}
\caption{Comparison of results in separated domains and united domains, respectively.}
\label{tab:ontonotes_together}
\end{table}

\subsection{Exp-II: Sequence labeling}
Secondly, we evaluate our model on sequence labeling task. It challenges the model to discover a shared architecture with hidden hierarchies to complete tasks that are different in semantic levels and significantly differ from each other. Different from Exp-I, to make the model more suitable for sequence labeling, we use a conditional random field (CRF)\citep{lafferty2001conditional} as the output layer.

\paragraph{Data}

We use OntoNotes 5.0 dataset\footnote{\url{https://catalog.ldc.upenn.edu/ldc2013t19}.} \citep{weischedel2013ontonotes} for our sequence labeling experiment.\\
OntoNotes contains several tasks across different domains, which helps us validate our model's ability to learn shared hierarchies. We select four tasks: part-of-speech tagging (POS tagging), chunking (Chunk), named entity recognition (NER), and semantic role labeling(SRL). The statistical details of the sub-datasets for each domain are shown in Table \ref{tab:ontonote}. There are four annotations for each sentence, corresponding to four tasks.


\begin{figure}[t]
   \centering \includegraphics[width=0.47\textwidth]{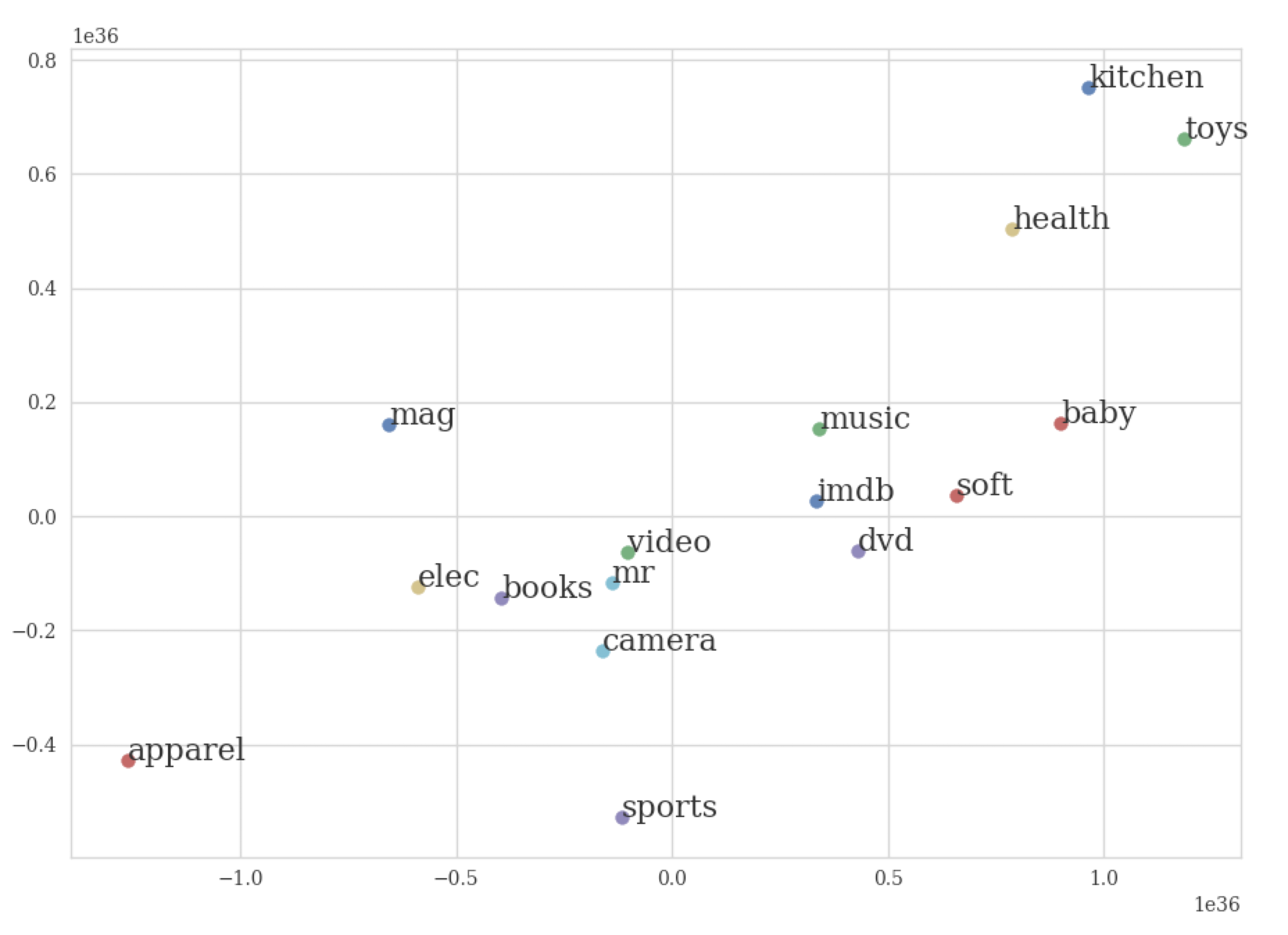}
  \caption{Task embedding for Exp-I (text classification), projected onto two-dimensional space by t-SNE. The full names of domains are displayed in Table \ref{tab:ontonote}.} \label{fig:task_embedding}
\end{figure}
\begin{figure*}[t]\small
\centering
\subfloat[The hierarchical architecture selected by the controller. Different colors represent different tasks. The lines and arrows show the chosen information flow of tasks. For example, given a sample of Chunking task, the controller sequentially chooses $M_1$, $M_2$, and $Stop$, forming the path in teal of the picture. The selective modules are originally disordered and are renamed $M_1 \cdots M_4$ for more explicit representation.]{
  \includegraphics[height=0.3\textheight, width=0.4\linewidth]{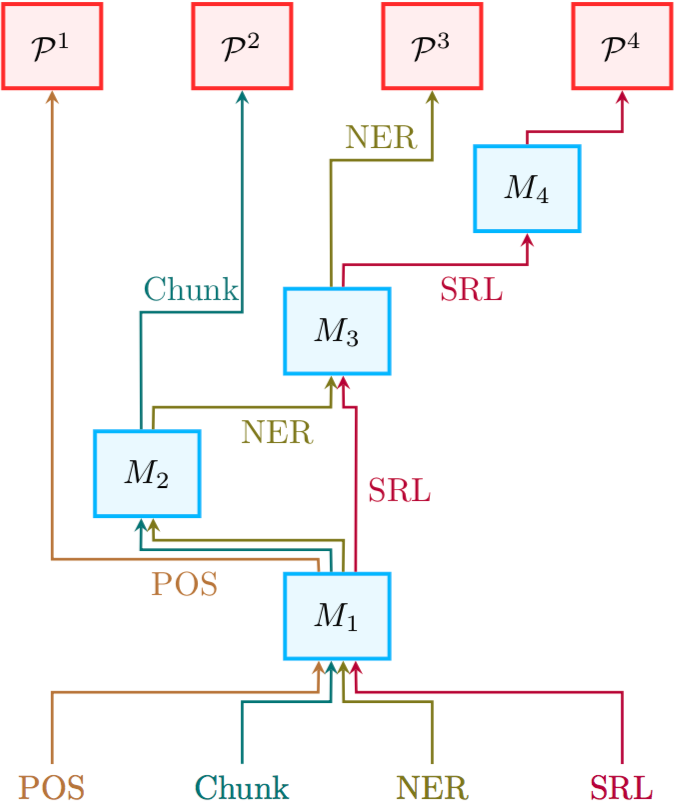}\label{fig:hierarchy}
  }
 \hspace{2em}
\subfloat[Probabilities of module selection at each step of the decision process. Probabilities are drawn from controller's module decision process after $\softmax$ function. The first four columns in every histogram represent the probability of choosing four different modules. The fifth column means the probability of deciding to $Stop$. For chunking, the decision sequence with the biggest probabilities is $M_1$, $M_2$, $Stop$, which corresponds with the path in \ref{fig:hierarchy}.]{
  \includegraphics[height=0.3\textheight, width=0.45\linewidth]{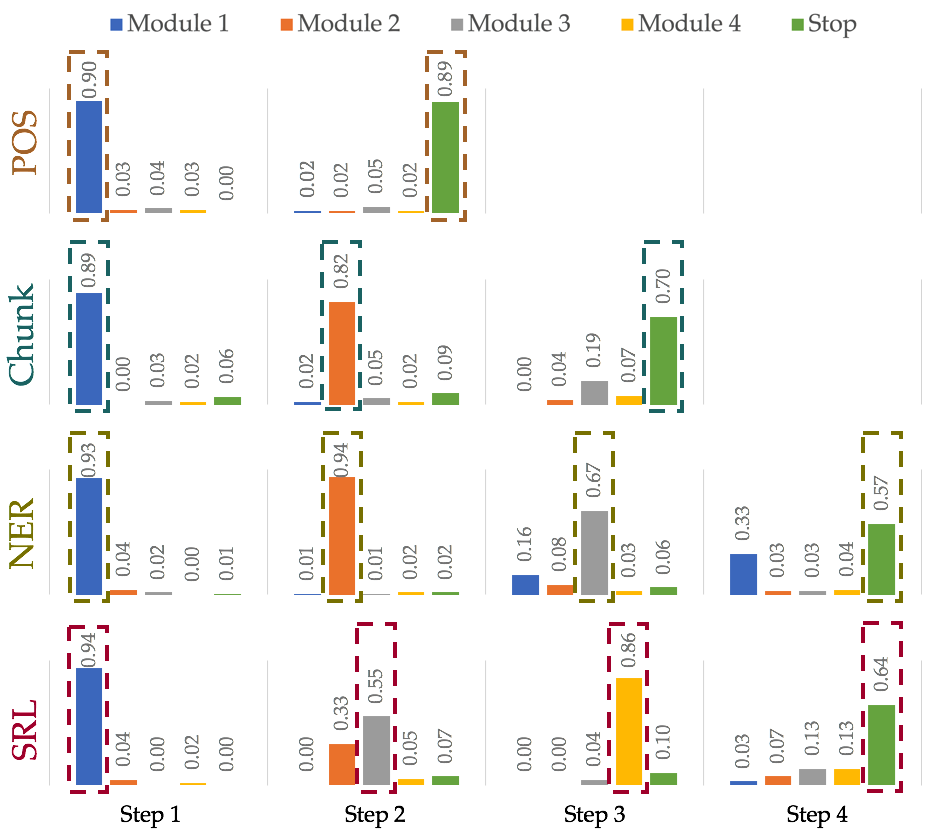}\label{fig:probs}
  }
  \hspace{0.5em}

  \caption{The selected architecture and the corresponding probabilities at every step of the decision process.}\label{fig:chunk_srl}
\end{figure*}

In contrast to Exp-I, where the differences among tasks are simply domains, tasks in this experiment differ from each other more significantly, and the relationships of tasks are more complicated. For example, solving NER problems can rarely help solving POS tagging task, while POS tagging may somehow provide low-level information for NER problems. There seems to be a hidden hierarchical structure of these tasks, and it challenges our model to find the hidden structure without any prior knowledge.
\paragraph{Result}

We regard every combination of (domain, problem) such as (bc, NER) as different tasks, accumulating to 7 * 4 - 1 = 27 tasks. We evaluate our model together with all baseline models mentioned in Exp-I, with a slight change of removing the average pooling classifier to every step in the sequence. Table \ref{tab:ontonotes_sep} shows the final accuracy of all these tasks. It is remarkable that our model achieves higher performance over all baseline models. It shows that our model successfully handles the situation of complicated task relationships.

Additionally, we carry out a small experiment of gathering data across domains together, reducing the number of tasks from 27 to 4. According to experiment results demonstrated in Table \ref{tab:ontonotes_together}, the performance with united domains is lower than that with separated domains. It shows that our model is capable of leveraging domain-specific knowledge.

\section{Analysis}

The high performance of our model suggests that the controller selects proper structures in different situations. To get a more intuitive understanding of how our model works, we explore and visualize the chosen architecture. Three main characteristics of our generated structure are as follows.

\subsection{Task Embedding Clustering}
In our text classification experiment, the structure generated by our model is capable of clustering tasks from similar domains together. To illustrate our clustering ability, we use t-SNE \citep{maaten2008visualizing} to project our task embedding onto two-dimensional space. The results are in Figure \ref{fig:task_embedding}.

From the illustration, we find that the task embedding shows the ability to cluster similar domains. For example, the task embedding of 'IMDB'(movie review) lies near 'video', 'music', and 'DVD'\textbf{}. It indicates that our model can distinguish characteristics of different tasks, and eventuall\textbf{}y finds a proper structure for the problem.

\begin{figure*}[t]\small
\centering
  \includegraphics[width=0.6\linewidth]{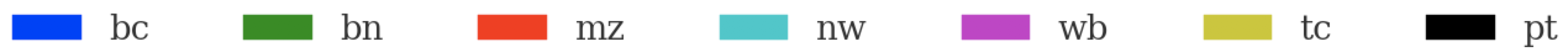}\label{fig:legends}
\\
\subfloat[3D view]{
  \includegraphics[width=0.275\linewidth]{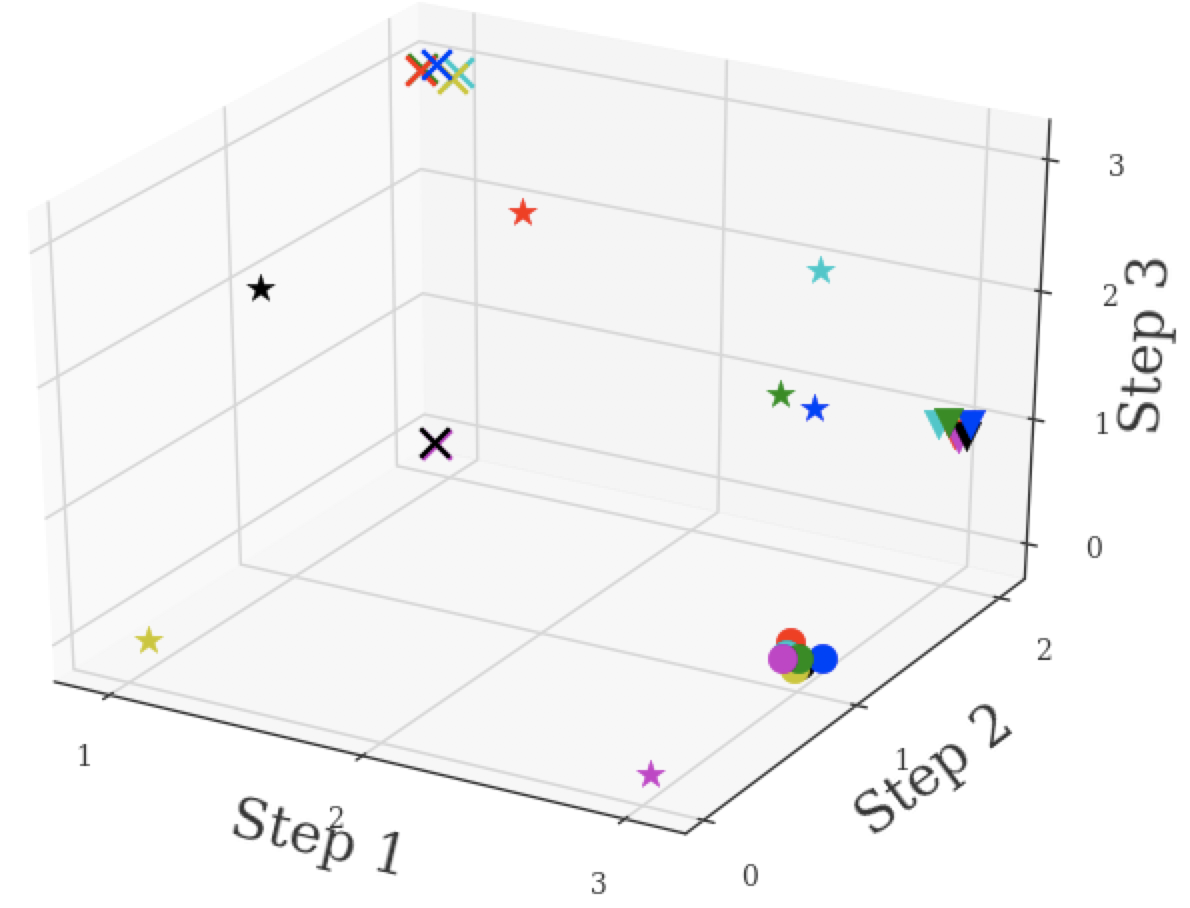}\label{fig:3d}
  }
\hspace{-0.8em}
\subfloat[Vertical view]{
  \includegraphics[width=0.235\linewidth]{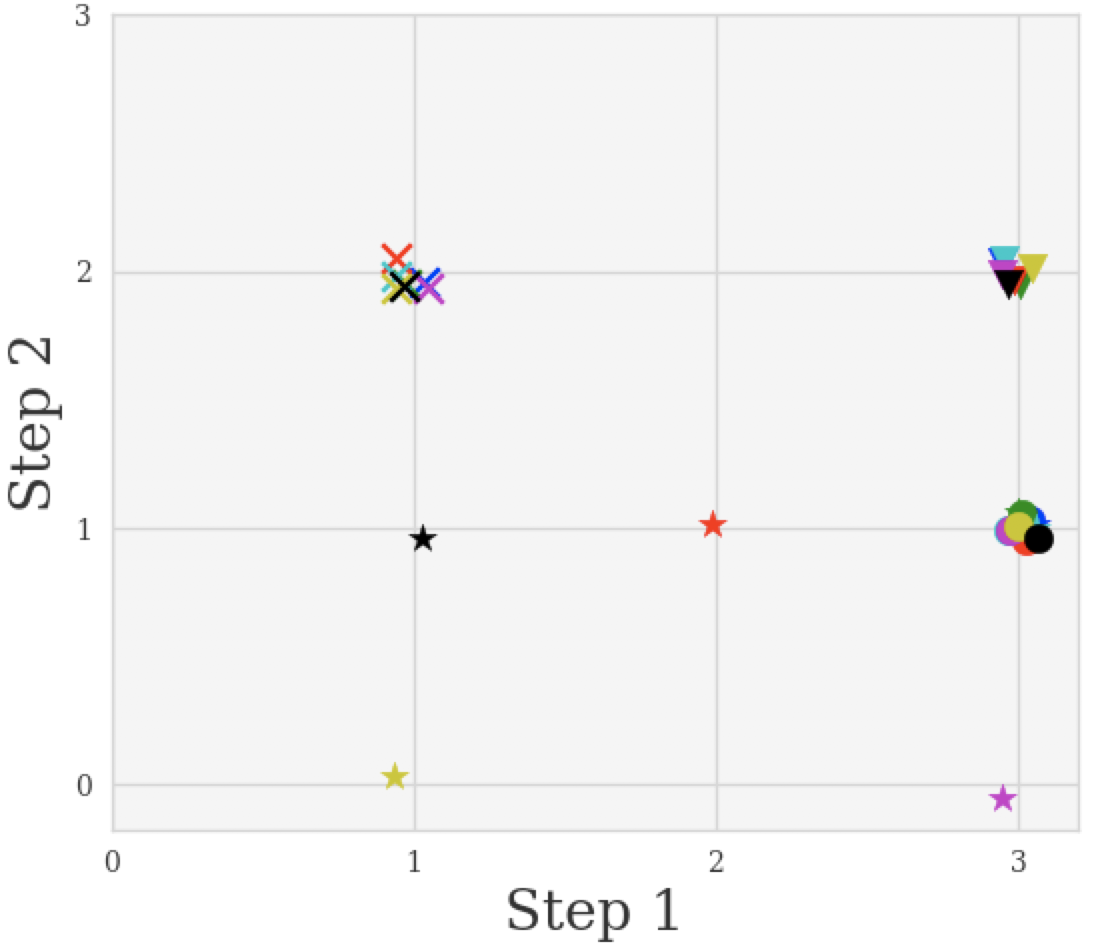}\label{fig:3dv}
  }
  \hspace{-0.8em}
\subfloat[Front view]{
  \includegraphics[width=0.235\linewidth]{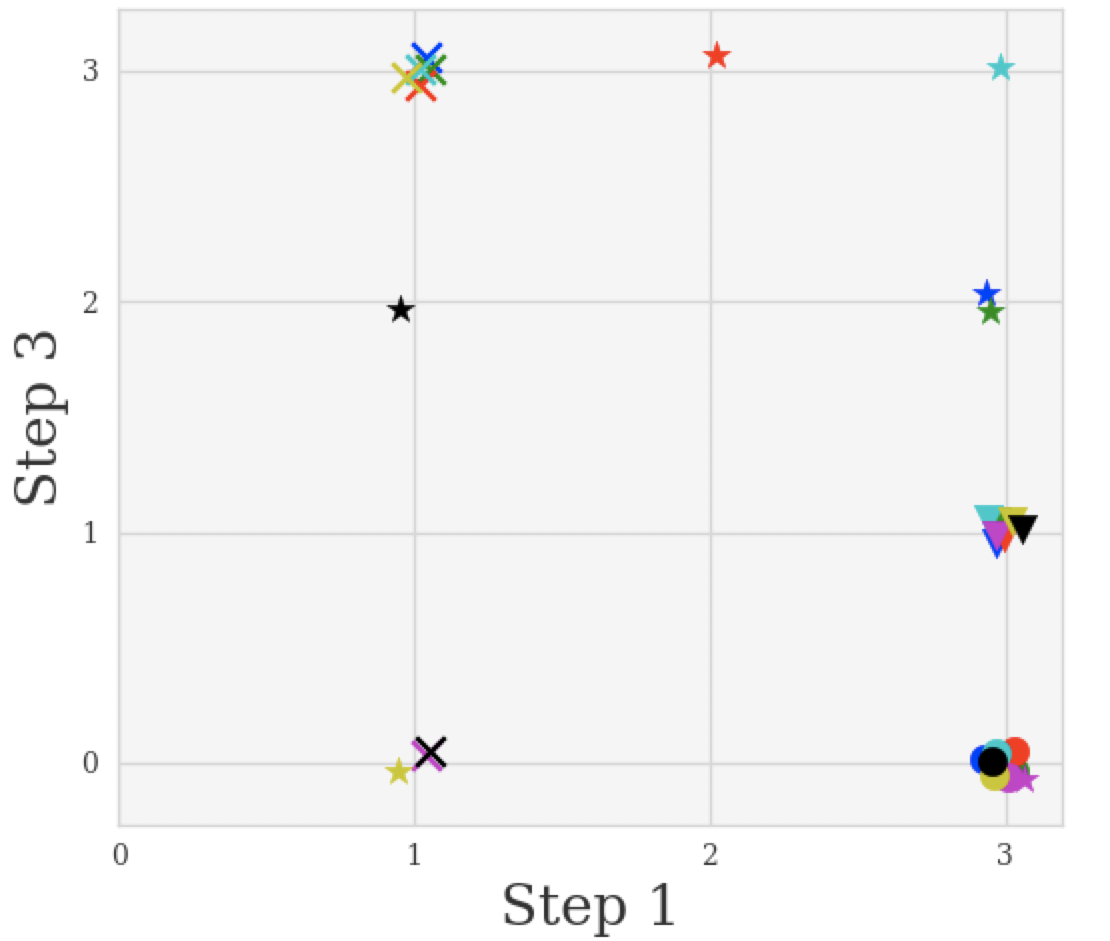}\label{fig:3df}
  }
  \hspace{-0.8em}
\subfloat[Left view]{
  \includegraphics[width=0.235\linewidth]{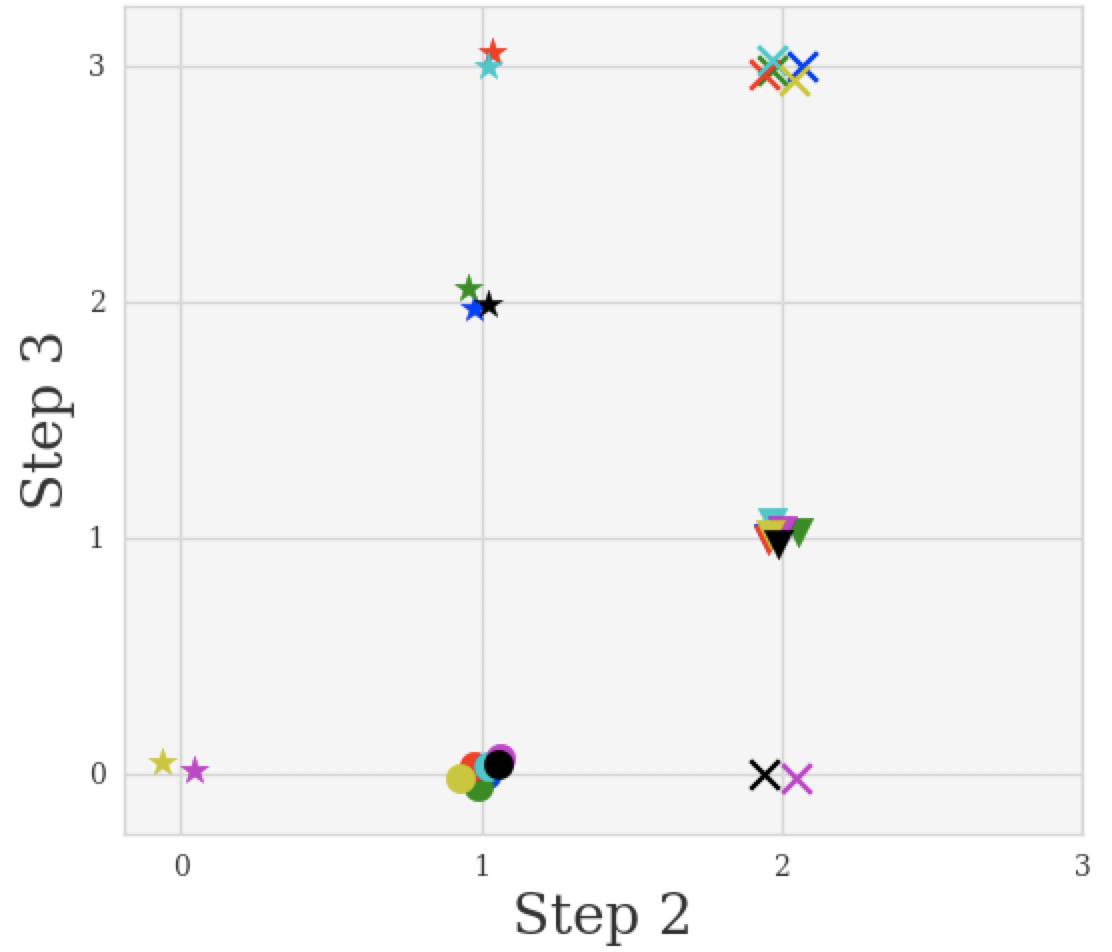}\label{fig:3ds}
  }
  \caption{Module selection sequence projected into three-dimensional space. Each point in the space represents a task. The position of it is determined by its module selection sequence. POS Tagging, Chunking, NER and SRL are respectively denoted by $\bullet$, $\blacktriangledown$, $\bigstar$, and \textbf{$\mathbf{\times}$}. Points align precisely on discrete positions. Slight shifts are imposed upon points for better visualization.}\label{fig:projection}
\end{figure*}

\subsection{Learned Hierarchical Architecture}

As proposed by \citep{hashimoto2017joint}, tasks of POS Tagging, Chunking, NER and sentence-level problems form a hierarchical linguistic structure, to which multi-task models should conform. To validate if our model learns a hierarchical architecture, we investigate the selected architecture in the 4-task experiment of sequence labeling. For each task, we visualize the module selection sequence of the biggest probability. The results are shown in Figure \ref{fig:hierarchy}.

In the beginning, the controller randomly chooses modules, trying structures that may contain loop paths or even self-loops. Nevertheless, our model eventually comes up with a hierarchical structure, with decision probabilities shown in Figure \ref{fig:probs}. It convincingly indicates that our model has learned to generate a superior hierarchical structure.

\subsection{Module Selection Clustering}

Another characteristic of the selected architecture is that similar tasks are clustered to share similar module structures. It can be illustrated if we regard the sequence of chosen modules as coordinates and project it into high-dimensional space. Since only three steps are used in consequence, Figure \ref{fig:projection} shows results in three-dimensional space.

There are some interesting patterns in the distributions of different tasks. Firstly, tasks of the same problem are likely to be neighboring. Secondly, high-level tasks like SRL are distributed more sparsely. Additionally, similar domains, for example, tasks in bc(broadcast conversation) have the same positions as tasks in bn(broadcast news). All these patterns found in the distributions show the ability of our model of clustering similar tasks and generating a reasonable structure.

\section{Related Work}

Multi-Task Learning has achieved superior performance in many applications \citep{collobert2008unified,glorot2011domain,liu2015representation,liu2016recurrent,duong2015low}. However, it is always done with hard or soft sharing of layers in the hope of obtaining a more generative representation shared between different tasks.

Because of restrictions on shared structure, so that only when facing the loosely relevant tasks or datasets, the effect will be good.
\citet{sogaard2016deep} and \citet{hashimoto2017joint} train their model considering linguistic hierarchies between related NLP tasks, by making lower level tasks
affect the lower levels of the representation. They show that given a hierarchy of tasks, the effectiveness of multitasking learning increases,  but they are pre-defined by human knowledge.

Recently, to break through the constraints of the shared structure, \citet{lu2017fully} proposes a bottom-up approach that dynamically creates branches of networks during training that promotes grouping of similar tasks. \citet{meyerson2017beyond} introduced as a method for learning how to apply layers in different ways at different depths for different tasks, while simultaneously learning the layers themselves. However it can only change the order of shared layers, cannot learn the hierarchies.

Other approaches related to ours include meta-learning and neural architecture search(NAS), meta-learning tries to learn a meta-model that generates a model for specific tasks. Most of them are using the meta-network to controls the parameters of the task-specific networks \citep{ravi2016optimization,chen2018metamtl}. The NAS is introduced in \citep{zoph2016neural,baker2016designing,pham2018efficient}, where it's applied to construct CNN for image classification. And \citet{liu2017hierarchical} has introduced a hierarchical genetic representation method into it. However, they are all dealing with single tasks, not the shared architecture search in Multi-Task Learning. \citet{rosenbaum2017routing}'s work is similar with us, but they have made restrictions on the optional position of modules. And they only optimize on the same type of tasks, like image classification, do not consider the latent hierarchy between tasks that enhance representation.

\section{Conclusion}
In this paper, we introduce a novel framework to explore the shared structure for MTL with automatic neural architecture search. Different from the existing MTL models, we use a controller network to select modules from a shared pool for individual tasks.
Through this way, we can learn an optimal shared structure for tasks to achieve better performance leverage the implicit relations between them. We evaluate our model on two types of tasks, have both achieved superior results. Besides, we have verified that our model can learn a reasonable hierarchy between tasks without human expertise.
\bibliographystyle{aaai}
\bibliography{nlp}

\end{document}